\pdfoutput=1

\documentclass[11pt]{article}

\usepackage[final]{EMNLP2023}

\usepackage{times}
\usepackage{latexsym}

\usepackage[demo]{graphicx}
\graphicspath{ {images/} }

\usepackage[T1]{fontenc}

\usepackage[utf8]{inputenc}

\usepackage{microtype}

\usepackage{inconsolata}

\usepackage{multirow}
\usepackage{tabularx}
\usepackage{booktabs}
\usepackage{makecell}
\usepackage{amsmath}

\usepackage{listings}
\usepackage{color}

\usepackage{paralist}

\definecolor{dkgreen}{rgb}{0,0.6,0}
\definecolor{gray}{rgb}{0.5,0.5,0.5}
\definecolor{mauve}{rgb}{0.58,0,0.82}

\def\FRMWRK/{LUNA}

\title{LUNA: A Framework for Language Understanding and Naturalness Assessment} 

\author{
    ~\textbf{Marat Saidov}\thanks{\ \ Equal contribution.}, 
    ~\textbf{Aleksandra Bakalova}$^*$,  
    ~\textbf{Ekaterina Taktasheva} \\ 
    ~\textbf{Vladislav Mikhailov},
    ~\textbf{Ekaterina Artemova}\\
    Work done at HSE University \\ 
    \small{
    \textbf{Correspondence:} \href{mailto:ekaterina.l.artemova@gmail.com}{\texttt{ekaterina.l.artemova@gmail.com}}
}}

\begin{document}
\maketitle
\begin{abstract}

The evaluation of Natural Language Generation (NLG) models has gained increased attention, urging the development of metrics that evaluate various aspects of generated text. \FRMWRK/ addresses this challenge by introducing a unified interface for 20 NLG evaluation metrics. These metrics are categorized based on their reference-dependence and the type of text representation they employ, from string-based n-gram overlap to the utilization of static embeddings and pre-trained language models.

The straightforward design of \FRMWRK/  allows for easy extension with novel metrics, requiring just a few lines of code. \FRMWRK/ offers a user-friendly tool for evaluating generated texts.

\end{abstract}

\section{Introduction}

In recent years, there has been a growing interest in finding better ways to evaluate how well natural language generation (NLG) models perform \cite{zhou-etal-2022-deconstructing,sai2022survey}. This involves creating new metrics that assess different aspects of the generated text, ranging from nuanced task-specific phenomena \cite{zhao-etal-2023-discoscore} to overall quality \cite{rei-etal-2020-comet} and its correlation with human judgements \cite{chhun-etal-2022-human}. However, comparing NLG models directly is becoming more challenging due to the increasing diversity of evaluation metrics and datasets used to showcase novel evaluation approaches. Nonetheless, a consistent and reproducible way to run evaluation is crucial for the progress in the field. 

Prior efforts have yielded valuable unified evaluation tools, such as HuggingFace Evaluate\footnote{\url{hf.co/evaluate/}}, which targets LLM performance, and \texttt{TextAttack} \cite{morris-etal-2020-textattack}, focused on large language models' (LLMs) robustness evaluation. Building on this progress, we introduce a specialized tool for evaluating NLG models with respect to a wide range of metrics.

In this paper, we present \FRMWRK/, a Python framework for language understanding and naturalness assessment. \FRMWRK/ offers a unified interface encompassing 20 NLG evaluation metrics widely used in popular sequence-to-sequence and open-ended text generation tasks. These metrics are categorized based on two primary design choices: \begin{inparaenum}[\itshape (i)] \item reference-based and reference-free metrics (i.e., metrics that require and do not require any reference to compare the generated output with), and 
\item the text representation they utilize, which range from string-based $n$-gram overlap to employing word embedding models and contextualized representations from pre-trained LLMs.  \end{inparaenum}  To streamline the evaluation, we provide two setups, in which either a single example or a batch of examples is evaluated. This ensures that  \FRMWRK/ can accommodate a wide range of evaluation scenarios, from fine-grained analysis to large-scale assessments.  \FRMWRK/ is agnostic to specific tasks and can be applied to assess the outputs of any generative LLMs. This includes task-specific models like machine translation, as well as more general open-ended generation models.

\FRMWRK/ leverages HuggingFace Transformers~\cite{wolf-etal-2020-transformers}, enabling  integration with any Transformer-based LLM required for specific metrics. Additionally, the framework's design allows for straightforward extensions with new metrics, requiring only a few lines of code. 

\FRMWRK/ is distributed under the MIT license as a Python package installed via direct cloning from GitHub\footnote{\url{https://github.com/Moonlight-Syntax/LUNA}}. Use cases for \FRMWRK/ include: \begin{inparaenum}[\itshape (i)] \item Comparative analysis and correlation of NLG evaluation metrics from existing literature;
    \item Ranking of generation hypotheses based on multiple metrics;
    \item Development and testing of novel NLG evaluation metrics. \end{inparaenum}

\begin{table*}[t!]
\centering
\resizebox{.8\textwidth}{!}{%
\begin{tabularx}{\textwidth}{p{3cm}|p{6cm}p{6cm}}
\toprule
 & \textbf{Reference-based}  & \textbf{Reference-free}  \\ \midrule
\multirowcell{6}{\textbf{String-based} \\ \textbf{metrics}} & BLEU \cite{papineni2002bleu} & Coverage \cite{grusky2018newsroom} \\
& ROUGE \cite{lin2004rouge} & Density \cite{grusky2018newsroom} \\
& chrF \cite{popovic2015chrf} & Compression \cite{grusky2018newsroom} \\
& METEOR  \cite{banerjee2005meteor} & Length \cite{fabbri2021summeval} \\ 
& \hfill & Novelty \cite{fabbri2021summeval} \\ 
& \hfill & Repetition \cite{fabbri2021summeval} \\ \hline 
\multirowcell{5}{\textbf{Embedding-based} \\ \textbf{metrics}} & ROUGE-We \cite{ng2015better}&  \\
& BERTScore \cite{zhangbertscore} & \\
& MoverScore \cite{zhao2019moverscore} & \\
& BaryScore \cite{colombo2021automatic} & \\
& DepthScore \cite{staerman2021pseudo} & \\ \midrule
\multirowcell{4}{\textbf{Model-based }\\ \textbf{metrics}} &  S3 \cite{peyrard2017learning} &  BLANC \cite{vasilyev2020fill} \\
& SummaQA \cite{scialom2019answers} & \\
& InfoLM \cite{colombo2022infolm} & \\
& \multicolumn{2}{c}{BARTScore \cite{yuan2021bartscore}} \\ 
\bottomrule
\end{tabularx}%
}
\caption{Metrics supported in the \FRMWRK/ framework.}
\label{table:all-metrics}
\end{table*}

\section{Framework}
\FRMWRK/ offers a unified and user-friendly interface for running a variety of NLG evaluation metrics. \autoref{table:all-metrics} lists NLG evaluation metrics currently supported by \FRMWRK/, which includes both traditional count-based metrics such as BLEU~\cite{papineni2002bleu} and ROUGE~\cite{lin2004rouge}, as well as newer metrics such as InfoLM~\cite{colombo2022infolm} and BARTScore~\cite{yuan2021bartscore}. The base class within \FRMWRK/ is implemented as  below:
\begin{lstlisting}[language=Python, caption=The base class in \FRMWRK/.]
class Metrics:
    def evaluate_batch(self, candidates: List[str], references: Optional[List[str]]) -> List[float]:
        *some code here*

    def evaluate_example(self, candidate: str, reference: Optional[str]) -> float:
        *some code here*
\end{lstlisting}

\paragraph{Overall design.} Each metric is implemented as a separate class, inheriting from the base class. The specific parameters for each metric are customized by providing them as input arguments to the initialization method. Evaluation can be performed in two ways: (i) using the \texttt{``evaluate\_example''} method to evaluate a single example, or (ii) using the \texttt{``evaluate\_batch''} method to evaluate a batch of examples.

In \FRMWRK/, evaluating the same input with multiple metrics is straightforward because all metrics utilize the same base interface. To accomplish this, the user simply needs to loop over all the required metrics, as demonstrated in the example below.

\begin{lstlisting}[language=Python, caption=Consequent run of two metrics.]
from luna.mover_score import MoverScoreMetrics
from luna.ngram import BLEUMetrics

metrics = [
    MoverScoreMetrics(
        n_gram=1,
        model_name="distilbert-base-uncased",
        compute_idfs=False,
        device="cuda"
    ),
    BLEUMetrics()
]
candidate = "Today is such a great day! I'm not sure if I want to go get some tea or just end it all."
reference = "What a good day today! Whether to go have a cup of tea or hang oneself."
results = [metric.evaluate_example(candidate, reference) for metric in metrics]
\end{lstlisting}

Additionally, we offer a specialized calculator mechanism that enables the simultaneous evaluation of multiple metrics on the same input with a single call. This mechanism also provides the option for parallel execution, where each metric runs in a separate process.

\paragraph{Input arguments.}  All metrics adhere to specific guidelines to ensure user-friendliness. During initialization, metric parameters are provided, including the underlying language model (via the model identifier in the HuggingFace hub), a customized list of stopwords, selected embeddings, and other relevant arguments. The extensive range of customizable parameters ensures that the metrics can be adapted to various models and languages, unless limited by metric architecture. Memory for storing language models or embeddings is also allocated during initialization.

\paragraph{Evaluation modes and tests.} To evaluate multiple candidates efficiently, we recommend using \texttt{``evaluate\_batch''} method, which is optimized for parallel processing. If a specific metric can be further optimized at its own level, the metric class overrides this method to implement the optimization.  Multiprocessing is employed, where such optimization is not feasible. This means that the \texttt{``evaluate\_example''} method is executed across multiple processes, taking advantage of parallel computing if multiple devices are available. In situations where optimization is not possible, calling ``evaluate\_batch'' is essentially the same as sequentially applying the \texttt{``evaluate\_example''} method to all candidates within a batch. To evaluate a single candidate without the need for parallel processing, we recommend using the \texttt{``evaluate\_example''} method. We provide tests for each metric and suggest that new users run these tests before using \FRMWRK/ to ensure that the framework functions correctly on their device.

\section{Components}

In \FRMWRK/, the NLG evaluation metrics are categorized into two coarse-grained categories: reference-based metrics and reference-free metrics, following the approach in \citet{chhun2022human}. \textit{Reference-based metrics} evaluate a candidate text by comparing it to a reference text, typically produced by human experts. \textit{Reference-free metrics}, on the other hand, rely solely on the candidate generation and, in some cases, on the input prompt. Both reference-free and reference-based metrics can be further divided into three fine-grained categories, depending on the type of information they access for evaluation: string-based, embedding-based, and model-based metrics. 

\paragraph{String-based metrics} String-based metrics refer to automatic metrics that operate on textual representations of the input, e.g. n-grams of words or characters. Such metrics, thus, rely only on surface level information, not taking semantics into account. Reference-based metrics of this type (e.g. BLEU \cite{papineni2002bleu}, ROUGE \cite{lin2004rouge}, inter alia) measure the word or character distance between the candidate generation and target text. Reference-free string-based metrics evaluate the degree of text overlap between the input and the model output, including novelty \cite{fabbri2021summeval}, text coverage, density, compression ratio, length, and repetition~\cite{grusky2018newsroom}.

\paragraph{Embedding-based metrics} Embedding-based metrics operate on word embeddings and use measures such as cosine similarity to evaluate the generated text against a reference. We include both, metrics relying on non-contextual static vectors, namely ROUGE-We \cite{ng2015better}, and metrics based on pretrained contextualized representations, such as BERTScore \cite{zhangbertscore} or MoverScore \cite{zhao2019moverscore}, among others. \FRMWRK/ can be utilized in many languages if a metric of interest supports a multilingual LM.

\paragraph{Model-based metrics} Model-based metrics include evaluation metrics, that employ regression or pre-trained language models to predict a score. We include learned metrics, such as S3 \cite{peyrard2017learning}, that are directly trained to predict human evaluation scores, as well as untrained metrics, for example, InfoLM \cite{colombo2022infolm} that uses information measures to predict the distance between two probability distributions without learning.



\subsection{Adding new components}

In the following section, we will elaborate on adding new components and extending \FRMWRK/. The metrics in \FRMWRK/ can be categorized based on their implementation into three groups:  \begin{inparaenum}[\itshape (i)]
  \item metrics with third-party library implementations;
  \item metrics with an implementation hosted in the reference repository; and
  \item metrics with an improved implementation.
\end{inparaenum}

\paragraph{Metrics with third-party library implementation} The complex logic needed to run the metric is already encapsulated within a third-party class or function. For instance, BLANC~\cite{vasilyev2020fill} utilizes its own PyPI package.

\begin{lstlisting}[language=Python, caption=Implementation of a third-party metric]
# BLANC (model-based, reference-free metric.)
from blanc import BlancHelp, BlancTune

class BlancMetrics(Metrics):
    def __init__(*args, **kwargs) -> None:
        # <...>
        if type == "help":
            self.metric = BlancHelp(*args, **kwargs)
        elif type == "tune":
            self.metrics = BlancTune(*args, **kwargs)
        else:
            raise ValueError(f"Type can be only help or tune. Got type = {type}")

\end{lstlisting}


To efficiently design the evaluation function for either one instance or a batch of instances, one should utilize the third-party implementation. This approach is far more effective than developing the inference process from scratch. Leveraging the existing third-party implementation re-uses the codebase that has been  thoroughly tested and optimized by the developers.

\paragraph{Metrics with an implementation hosted in the reference repository}
Metrics with an implementation hosted in the reference repository are saved in the \texttt{``luna.sources``} and then imported during the initialization of the corresponding metric. It is important to note that this approach adds some overhead to the codebase, which we address in the next section. Here is an example of the BaryScore metric~\cite{colombo2021automatic} pseudo code that uses the \texttt{``luna.sources``} module:

\begin{lstlisting}[language=Python, caption=Implementation of a metric adopted from a reference repository.]
from luna.base import Metrics
from luna.sources.bary_score import BaryScoreMetricsInternal

class BaryScoreMetrics(Metrics):
    *some code here*
    def __init__(self, *args, **kwargs) -> None:
        self.bary_score_metrics = BaryScoreMetricsInternal(*args, **kwargs)

    def _run_bary_score(self, hyps: tp.List[str], refs: tp.List[str]) -> tp.Dict[str, tp.Any]:
        idf_dict_hyp, idf_dict_ref = (
            self.bary_score_metrics
            .prepare_idfs(hyps, refs)
        )
        metrics_dict = (
            self.bary_score_metrics
            .evaluate_batch(
                hyps, refs, idf_dict_hyp, idf_dict_ref
            )
        )
        return metrics_dict

    def evaluate_example(self, hyp: str, ref: str) -> float:
        metrics_dict = self._run_bary_score([hyp], [ref])
        return metrics_dict[
                        self.bary_score_key
                        ][0]

    def evaluate_batch(self, hyps: tp.List[str], refs: tp.List[str]) -> tp.List[float]:
        *similar code goes here*

\end{lstlisting}

\paragraph{Metrics with an improved implementation} For this class the available implementation is used but significantly improved. In particular, we optimize the SummaQA metric~\cite{scialom2019answers} for both Russian and English languages since for each language different question generators and answer predictors are required. Besdies, the algorithm which averages the answer correctness across masked sentences is precisely tested on the mocked dataset. Finally, due to the right decomposition we unify all of the metrics to the specific interface which allows us to orchestrate them and to calculate them simultaneously on a dataset of interest. Nevertheless, it is important to reduce the computational costs on the client-side and to move the most computationally intensive parts of the framework to the library- or server-side.

\subsection{Enhancing \FRMWRK/ to HuggingFace Spaces}


Several metrics in \FRMWRK/ rely on Transformer-based LLMs. While \FRMWRK/ currently operates exclusively in local environments, we are developing an extension to facilitate remote endpoints. For deploying the \FRMWRK/ framework, HuggingFace Spaces serves as a free option for hosting metrics. Take SummaQA, BaryScore, or DepthScore as examples. These computationally-heavy metrics could be treated as remote containers with public inference endpoints, which \FRMWRK/ accesses through a Python API.


\section{Evaluating NLG with \FRMWRK/}

Here we provide examples of how to use \FRMWRK/. Note that there is a specific additional \texttt{calculator} mechanism, which allows metrics to be processed simultaneously in one single call. It provides the parallel execution option.

\begin{lstlisting}
from luna.calculate import Calculator

# candidates and references are batches
calculator = Calculator(execute_parallel=True)
metrics_dict = calculator.calculate(
    metrics=[depth_score_metrics, s3_metrics],
    candidates=candidates,
    references=references
)

print(metrics_dict)
>>> {"DepthScore": ..., "S3": ...}
\end{lstlisting}


\subsection{Reference-free usage}

Some evaluations in \FRMWRK/ do not require references. These metrics operate directly on raw hypotheses and are contained in a separate module called \texttt{``luna.reference\_free''}. Unlike other, metrics in this category do not rely on pre-trained LLMs. Instead, they perform basic counting operations on sentences, such as measuring {\it length}, {\it compression}, or {\it repetition}.

The arguments for the string-based reference-free metrics are uniform across all metrics in the \texttt{``luna.reference\_free''} module, as they are inherited from the parent class. Here is an example of a string-based reference-free metric:

\begin{lstlisting}
from luna.calculate import Calculator
from luna.reference_free import Length, Compression, Repetition

rf_args = {"n_gram": 3, "tokenize": True}
length_metrics = LengthMetrics(**rf_args)
compression_metrics = CompressionMetrics(**rf_args)
repetition_metrics = RepetitionMetrics(**rf_args)

# candidates is a batch, references is None
calculator = Calculator(execute_parallel=True)
metrics_dict = calculator.calculate(
    metrics=[length_metrics, compression_metrics, repetition_metrics],
    candidates=[...],
    references=None
)

print(metrics_dict)
>>> {"Length": ..., "Compression": ..., "Repetition": ...}

\end{lstlisting}

However, there are exceptions, which rely on embeddings and pre-trained LLMs. BLANC is an example of a reference-free metric, but it is model-based and typically requires a GPU for computation.

\subsection{Sentence-level usage}

The majority of evaluations in \FRMWRK/ are designed for sentence-level usage. Below is an example demonstrating the initialization of classes and computation of metrics at the sentence level. This example follows the default usage pattern of \FRMWRK/.

\begin{lstlisting}
from luna.rouge_we import RougeWeMetrics

candidate = "Today is such a great day! I'm not sure if I want to go get some tea or just end it all."
reference = "What a good day today! Whether to go have a cup of tea or hang oneself."

rouge_we_metrics = RougeWeMetrics(**rouge_we_kwargs)
rouge_value = rouge_we_metrics.evaluate_example(
    hyp=candidate, 
    ref=reference
    )

print(isinstance(rouge_value, float))
>>> True
\end{lstlisting}

Note that the \texttt{``evaluate\_example''} method returns a single floating-point number, while the \texttt{``evaluate\_batch''} method outputs a list of floating-point numbers.

\subsection{Corpus-level usage}

Within \FRMWRK/, evaluations are designed to indirectly aggregate corpus-level statistics. For instance, some metrics calculate the TF-IDF matrix over the entire corpus. However, there is also a dedicated corpus-level evaluation metric, namely SummaQA.

Processing batches and the entire corpus are conceptually similar situations. The main difference is the metric's method name and any exceptions or warnings that may be triggered. For example, let us consider SummaQA:

\begin{lstlisting}
from luna.summaqa import SummaQAMetrics

# score_to_return is either "fscore" or "prob"
summaqa = SummaQAMetrics(lang="ru", score_to_return="fscore")

# Evaluate the single sentence
summaqa.evaluate_example(...)
>>> RuntimeError: Separate examples evaluation is not supported for corpus-level metrics 

# Evaluate the batch
summaqa.evaluate_batch(...)
>>> Warning: Batch processing is considered as processing the textual corpus

# Evaluate the corpus
corpus = ["Very first sentence.", "Continuing.", "The end of the large corpus"]
summarization_candidates = ["First.", "Middle.", "Last."]
summaqa_metrics = summaqa.evaluate_corpus(
    candidates=summarization_candidates,
    references=corpus
)

print(len(summaqa_metrics) == len(corpus))
>>> True

\end{lstlisting}

\section{Related work}

We adopt the established approach of HuggingFace Transformers in developing a unified interface for NLP metrics. \FRMWRK/  integrates with HuggingFace Transformers, leveraging LLMs from the Transformer Model Hub.  

Prior efforts in developing NLG evaluation led to creation of multiple tools. NLG-eval \cite{sharma2017relevance} focuses mainly on string-based and embedding-based metrics and does not support learnable model-based evaluators. Jury \cite{obss2021jury} offers a combination of numerical and text metrics that can be used in a broad range of NLP tasks, of which 10 are aimed at NLG evaluation. TorchMetrics \cite{detlefsen2022torchmetrics} offers a large collection of machine learning metrics implemented with PyTorch, including 16 tailored for language tasks, and among them, seven implemented specifically to NLG evaluation. The HuggingFace Evaluation framework is a versatile tool designed to evaluate a wide range of NLP tasks without a specific domain focus. However, it may not include some of the most recent NLG evaluation metrics. \FRMWRK/ surpasses these tools by providing a more extensive set of NLG evaluation metrics and further advancements. 

A concurrent work by \citet{frisoni-etal-2022-nlg} introduces the NLG-Metricverse framework. While there is a significant overlap between the metrics supported by NLG-Metricverse and \FRMWRK/, it is worth noting that \FRMWRK/ places a stronger emphasis on efficient data processing and supports batch-wise evaluation unlike NLG-Metricverse.

\paragraph{Research questions in NLG evaluation.} Although frameworks like \FRMWRK/ provide a useful tool for a researchers and developers, they fall short in addressing the deficiencies of current NLG evaluation approaches. Several works made attempts to explore and improve the alignment between automated metrics and human judgment \cite{caglayan-etal-2020-curious, hanna-bojar-2021-fine}, finding that current metrics not only show weak correlation with human scores \cite{zhang-etal-2004-interpreting, novikova-etal-2017-need}, but exhibit social biases \cite{gao2022social} and system preferences \cite{callison-burch-etal-2006-evaluating}. Other works focus on meta-evaluation of automatic NLG metrics \cite{nimah-etal-2023-nlg, sai-etal-2021-perturbation}.

\section{Future work} The development of \FRMWRK/ is ongoing, with several potential directions for expansion. Firstly, there are plans to incorporate \textbf{additional NLG evaluation metrics}, including SUPERT \cite{gao-etal-2020-supert}, an example of reference-free embedding-based metric, as well as more recent metrics like RoMe \cite{rony-etal-2022-rome}, and DiscoScore \cite{zhao-etal-2023-discoscore}. 

Secondly, we plan to add support for \textbf{multi-reference formats}, similar to what has been implemented in \cite{obss2021jury}. This will enable us to relax the strict exact match constraint that currently dominates event embedding-based metrics and account for structural variability \cite{fomicheva-etal-2020-multi}.

Thirdly, we will add an \textbf{explicit mechanism to rank generation hypotheses} based on either a single metric or multiple metrics, according to multi-criteria decision-making rules, as discussed by \citet{colombo2022best}.

Finally, we can achieve a significant speedup in runtime by \textbf{enhancing support for multiple devices} and implementing more efficient parallelization techniques.

\section{Conclusion}
In this paper, we introduced \FRMWRK/ — a framework for evaluating natural language generation. It provides an easy-to-use interface to employ 20 popular NLG metrics varying in their type. \FRMWRK/ supports efficient computation of the metric scores at both instance and batch level, and can be easily extended with new metrics of interest. We hope that \FRMWRK/ provides NLP practitioners with an accessible and comprehensive tool for evaluating NLG models and assessing their performance.

\bibliography{anthology,custom}

\begin{thebibliography}{40}
\expandafter\ifx\csname natexlab\endcsname\relax\def\natexlab#1{#1}\fi

\bibitem[{Banerjee and Lavie(2005)}]{banerjee2005meteor}
Satanjeev Banerjee and Alon Lavie. 2005.
\newblock Meteor: An automatic metric for mt evaluation with improved
  correlation with human judgments.
\newblock In \emph{Proceedings of the acl workshop on intrinsic and extrinsic
  evaluation measures for machine translation and/or summarization}, pages
  65--72.

\bibitem[{Caglayan et~al.(2020)Caglayan, Madhyastha, and
  Specia}]{caglayan-etal-2020-curious}
Ozan Caglayan, Pranava Madhyastha, and Lucia Specia. 2020.
\newblock \href {https://doi.org/10.18653/v1/2020.coling-main.210} {Curious
  case of language generation evaluation metrics: A cautionary tale}.
\newblock In \emph{Proceedings of the 28th International Conference on
  Computational Linguistics}, pages 2322--2328, Barcelona, Spain (Online).
  International Committee on Computational Linguistics.

\bibitem[{Callison-Burch et~al.(2006)Callison-Burch, Osborne, and
  Koehn}]{callison-burch-etal-2006-evaluating}
Chris Callison-Burch, Miles Osborne, and Philipp Koehn. 2006.
\newblock \href {https://aclanthology.org/E06-1032} {Re-evaluating the role of
  {B}leu in machine translation research}.
\newblock In \emph{11th Conference of the {E}uropean Chapter of the Association
  for Computational Linguistics}, pages 249--256, Trento, Italy. Association
  for Computational Linguistics.

\bibitem[{Cavusoglu et~al.(2022)Cavusoglu, Akyon, Sert, and
  Cengiz}]{obss2021jury}
Devrim Cavusoglu, Fatih~Cagatay Akyon, Ulas Sert, and Cemil Cengiz. 2022.
\newblock \href {https://doi.org/10.5281/zenodo.6108229} {{Jury: Comprehensive
  NLP Evaluation toolkit}}.

\bibitem[{Chhun et~al.(2022{\natexlab{a}})Chhun, Colombo, Suchanek, and
  Clavel}]{chhun-etal-2022-human}
Cyril Chhun, Pierre Colombo, Fabian~M. Suchanek, and Chlo{\'e} Clavel.
  2022{\natexlab{a}}.
\newblock \href {https://aclanthology.org/2022.coling-1.509} {Of human criteria
  and automatic metrics: A benchmark of the evaluation of story generation}.
\newblock In \emph{Proceedings of the 29th International Conference on
  Computational Linguistics}, pages 5794--5836, Gyeongju, Republic of Korea.
  International Committee on Computational Linguistics.

\bibitem[{Chhun et~al.(2022{\natexlab{b}})Chhun, Colombo, Suchanek, and
  Clavel}]{chhun2022human}
Cyril Chhun, Pierre Colombo, Fabian~M Suchanek, and Chlo{\'e} Clavel.
  2022{\natexlab{b}}.
\newblock Of human criteria and automatic metrics: A benchmark of the
  evaluation of story generation.
\newblock In \emph{29th International Conference on Computational Linguistics
  (COLING 2022)}.

\bibitem[{Colombo et~al.(2022{\natexlab{a}})Colombo, Noiry, Irurozki, and
  Cl{\'e}men{\c{c}}on}]{colombo2022best}
Pierre Colombo, Nathan Noiry, Ekhine Irurozki, and St{\'e}phan
  Cl{\'e}men{\c{c}}on. 2022{\natexlab{a}}.
\newblock What are the best systems? new perspectives on nlp benchmarking.
\newblock \emph{Advances in Neural Information Processing Systems},
  35:26915--26932.

\bibitem[{Colombo et~al.(2021)Colombo, Staerman, Clavel, and
  Piantanida}]{colombo2021automatic}
Pierre Colombo, Guillaume Staerman, Chlo{\'e} Clavel, and Pablo Piantanida.
  2021.
\newblock Automatic text evaluation through the lens of wasserstein
  barycenters.
\newblock In \emph{Proceedings of the 2021 Conference on Empirical Methods in
  Natural Language Processing}, pages 10450--10466.

\bibitem[{Colombo et~al.(2022{\natexlab{b}})Colombo, Clavel, and
  Piantanida}]{colombo2022infolm}
Pierre Jean~A Colombo, Chlo{\'e} Clavel, and Pablo Piantanida.
  2022{\natexlab{b}}.
\newblock Infolm: A new metric to evaluate summarization \& data2text
  generation.
\newblock In \emph{Proceedings of the AAAI Conference on Artificial
  Intelligence}, volume~36, pages 10554--10562.

\bibitem[{Detlefsen et~al.(2022)Detlefsen, Borovec, Schock, Jha, Koker,
  Di~Liello, Stancl, Quan, Grechkin, and Falcon}]{detlefsen2022torchmetrics}
Nicki~Skafte Detlefsen, Jiri Borovec, Justus Schock, Ananya~Harsh Jha, Teddy
  Koker, Luca Di~Liello, Daniel Stancl, Changsheng Quan, Maxim Grechkin, and
  William Falcon. 2022.
\newblock Torchmetrics-measuring reproducibility in pytorch.
\newblock \emph{Journal of Open Source Software}, 7(70):4101.

\bibitem[{Fabbri et~al.(2021)Fabbri, Kryściński, McCann, Xiong, Socher, and
  Radev}]{fabbri2021summeval}
Alexander~R. Fabbri, Wojciech Kryściński, Bryan McCann, Caiming Xiong,
  Richard Socher, and Dragomir Radev. 2021.
\newblock \href {http://arxiv.org/abs/2007.12626} {Summeval: Re-evaluating
  summarization evaluation}.

\bibitem[{Fomicheva et~al.(2020)Fomicheva, Specia, and
  Guzm{\'a}n}]{fomicheva-etal-2020-multi}
Marina Fomicheva, Lucia Specia, and Francisco Guzm{\'a}n. 2020.
\newblock \href {https://doi.org/10.18653/v1/2020.acl-main.113}
  {Multi-hypothesis machine translation evaluation}.
\newblock In \emph{Proceedings of the 58th Annual Meeting of the Association
  for Computational Linguistics}, pages 1218--1232, Online. Association for
  Computational Linguistics.

\bibitem[{Frisoni et~al.(2022)Frisoni, Carbonaro, Moro, Zammarchi, and
  Avagnano}]{frisoni-etal-2022-nlg}
Giacomo Frisoni, Antonella Carbonaro, Gianluca Moro, Andrea Zammarchi, and
  Marco Avagnano. 2022.
\newblock \href {https://aclanthology.org/2022.coling-1.306}
  {{NLG}-metricverse: An end-to-end library for evaluating natural language
  generation}.
\newblock In \emph{Proceedings of the 29th International Conference on
  Computational Linguistics}, pages 3465--3479, Gyeongju, Republic of Korea.
  International Committee on Computational Linguistics.

\bibitem[{Gao and Wan(2022)}]{gao2022social}
Mingqi Gao and Xiaojun Wan. 2022.
\newblock \href {http://arxiv.org/abs/2210.08859} {Social biases in automatic
  evaluation metrics for nlg}.

\bibitem[{Gao et~al.(2020)Gao, Zhao, and Eger}]{gao-etal-2020-supert}
Yang Gao, Wei Zhao, and Steffen Eger. 2020.
\newblock \href {https://doi.org/10.18653/v1/2020.acl-main.124} {{SUPERT}:
  Towards new frontiers in unsupervised evaluation metrics for multi-document
  summarization}.
\newblock In \emph{Proceedings of the 58th Annual Meeting of the Association
  for Computational Linguistics}, pages 1347--1354, Online. Association for
  Computational Linguistics.

\bibitem[{Grusky et~al.(2018)Grusky, Naaman, and Artzi}]{grusky2018newsroom}
Max Grusky, Mor Naaman, and Yoav Artzi. 2018.
\newblock Newsroom: A dataset of 1.3 million summaries with diverse extractive
  strategies.
\newblock In \emph{Proceedings of the 2018 Conference of the North American
  Chapter of the Association for Computational Linguistics: Human Language
  Technologies, Volume 1 (Long Papers)}, pages 708--719.

\bibitem[{Hanna and Bojar(2021)}]{hanna-bojar-2021-fine}
Michael Hanna and Ond{\v{r}}ej Bojar. 2021.
\newblock \href {https://aclanthology.org/2021.wmt-1.59} {A fine-grained
  analysis of {BERTS}core}.
\newblock In \emph{Proceedings of the Sixth Conference on Machine Translation},
  pages 507--517, Online. Association for Computational Linguistics.

\bibitem[{Lin(2004)}]{lin2004rouge}
Chin-Yew Lin. 2004.
\newblock Rouge: A package for automatic evaluation of summaries.
\newblock In \emph{Text summarization branches out}, pages 74--81.

\bibitem[{Morris et~al.(2020)Morris, Lifland, Yoo, Grigsby, Jin, and
  Qi}]{morris-etal-2020-textattack}
John Morris, Eli Lifland, Jin~Yong Yoo, Jake Grigsby, Di~Jin, and Yanjun Qi.
  2020.
\newblock \href {https://doi.org/10.18653/v1/2020.emnlp-demos.16}
  {{T}ext{A}ttack: A framework for adversarial attacks, data augmentation, and
  adversarial training in {NLP}}.
\newblock In \emph{Proceedings of the 2020 Conference on Empirical Methods in
  Natural Language Processing: System Demonstrations}, pages 119--126, Online.
  Association for Computational Linguistics.

\bibitem[{Ng and Abrecht(2015)}]{ng2015better}
Jun~Ping Ng and Viktoria Abrecht. 2015.
\newblock Better summarization evaluation with word embeddings for rouge.
\newblock In \emph{Proceedings of the 2015 Conference on Empirical Methods in
  Natural Language Processing}, pages 1925--1930.

\bibitem[{Nimah et~al.(2023)Nimah, Fang, Menkovski, and
  Pechenizkiy}]{nimah-etal-2023-nlg}
Iftitahu Nimah, Meng Fang, Vlado Menkovski, and Mykola Pechenizkiy. 2023.
\newblock \href {https://doi.org/10.18653/v1/2023.acl-long.69} {{NLG}
  evaluation metrics beyond correlation analysis: An empirical metric
  preference checklist}.
\newblock In \emph{Proceedings of the 61st Annual Meeting of the Association
  for Computational Linguistics (Volume 1: Long Papers)}, pages 1240--1266,
  Toronto, Canada. Association for Computational Linguistics.

\bibitem[{Novikova et~al.(2017)Novikova, Du{\v{s}}ek, Cercas~Curry, and
  Rieser}]{novikova-etal-2017-need}
Jekaterina Novikova, Ond{\v{r}}ej Du{\v{s}}ek, Amanda Cercas~Curry, and Verena
  Rieser. 2017.
\newblock \href {https://doi.org/10.18653/v1/D17-1238} {Why we need new
  evaluation metrics for {NLG}}.
\newblock In \emph{Proceedings of the 2017 Conference on Empirical Methods in
  Natural Language Processing}, pages 2241--2252, Copenhagen, Denmark.
  Association for Computational Linguistics.

\bibitem[{Papineni et~al.(2002)Papineni, Roukos, Ward, and
  Zhu}]{papineni2002bleu}
Kishore Papineni, Salim Roukos, Todd Ward, and Wei-Jing Zhu. 2002.
\newblock Bleu: a method for automatic evaluation of machine translation.
\newblock In \emph{Proceedings of the 40th annual meeting of the Association
  for Computational Linguistics}, pages 311--318.

\bibitem[{Peyrard et~al.(2017)Peyrard, Botschen, and
  Gurevych}]{peyrard2017learning}
Maxime Peyrard, Teresa Botschen, and Iryna Gurevych. 2017.
\newblock Learning to score system summaries for better content selection
  evaluation.
\newblock In \emph{Proceedings of the Workshop on New Frontiers in
  Summarization}, pages 74--84.

\bibitem[{Popovi{\'c}(2015)}]{popovic2015chrf}
Maja Popovi{\'c}. 2015.
\newblock chrf: character n-gram f-score for automatic mt evaluation.
\newblock In \emph{Proceedings of the tenth workshop on statistical machine
  translation}, pages 392--395.

\bibitem[{Rei et~al.(2020)Rei, Stewart, Farinha, and
  Lavie}]{rei-etal-2020-comet}
Ricardo Rei, Craig Stewart, Ana~C Farinha, and Alon Lavie. 2020.
\newblock \href {https://doi.org/10.18653/v1/2020.emnlp-main.213} {{COMET}: A
  neural framework for {MT} evaluation}.
\newblock In \emph{Proceedings of the 2020 Conference on Empirical Methods in
  Natural Language Processing (EMNLP)}, pages 2685--2702, Online. Association
  for Computational Linguistics.

\bibitem[{Rony et~al.(2022)Rony, Kovriguina, Chaudhuri, Usbeck, and
  Lehmann}]{rony-etal-2022-rome}
Md~Rashad Al~Hasan Rony, Liubov Kovriguina, Debanjan Chaudhuri, Ricardo Usbeck,
  and Jens Lehmann. 2022.
\newblock \href {https://doi.org/10.18653/v1/2022.acl-long.387} {{R}o{M}e: A
  robust metric for evaluating natural language generation}.
\newblock In \emph{Proceedings of the 60th Annual Meeting of the Association
  for Computational Linguistics (Volume 1: Long Papers)}, pages 5645--5657,
  Dublin, Ireland. Association for Computational Linguistics.

\bibitem[{Sai et~al.(2021)Sai, Dixit, Sheth, Mohan, and
  Khapra}]{sai-etal-2021-perturbation}
Ananya~B. Sai, Tanay Dixit, Dev~Yashpal Sheth, Sreyas Mohan, and Mitesh~M.
  Khapra. 2021.
\newblock \href {https://doi.org/10.18653/v1/2021.emnlp-main.575} {Perturbation
  {C}heck{L}ists for evaluating {NLG} evaluation metrics}.
\newblock In \emph{Proceedings of the 2021 Conference on Empirical Methods in
  Natural Language Processing}, pages 7219--7234, Online and Punta Cana,
  Dominican Republic. Association for Computational Linguistics.

\bibitem[{Sai et~al.(2022)Sai, Mohankumar, and Khapra}]{sai2022survey}
Ananya~B Sai, Akash~Kumar Mohankumar, and Mitesh~M Khapra. 2022.
\newblock {A survey of evaluation metrics used for NLG systems}.
\newblock \emph{ACM Computing Surveys (CSUR)}, 55(2):1--39.

\bibitem[{Scialom et~al.(2019)Scialom, Lamprier, Piwowarski, and
  Staiano}]{scialom2019answers}
Thomas Scialom, Sylvain Lamprier, Benjamin Piwowarski, and Jacopo Staiano.
  2019.
\newblock Answers unite! unsupervised metrics for reinforced summarization
  models.
\newblock In \emph{2019 Conference on Empirical Methods in Natural Language
  Processing and the 9th International Joint Conference on Natural Language
  Processing (EMNLP-IJCNLP)}, pages 3237--3247. Association for Computational
  Linguistics.

\bibitem[{Sharma et~al.(2017)Sharma, Asri, Schulz, and
  Zumer}]{sharma2017relevance}
Shikhar Sharma, Layla~El Asri, Hannes Schulz, and Jeremie Zumer. 2017.
\newblock Relevance of unsupervised metrics in task-oriented dialogue for
  evaluating natural language generation.
\newblock \emph{arXiv preprint arXiv:1706.09799}.

\bibitem[{Staerman et~al.(2021)Staerman, Mozharovskyi, Colombo,
  Cl{\'e}men{\c{c}}on, and d'Alch{\'e} Buc}]{staerman2021pseudo}
Guillaume Staerman, Pavlo Mozharovskyi, Pierre Colombo, St{\'e}phan
  Cl{\'e}men{\c{c}}on, and Florence d'Alch{\'e} Buc. 2021.
\newblock A pseudo-metric between probability distributions based on
  depth-trimmed regions.
\newblock \emph{arXiv preprint arXiv:2103.12711}.

\bibitem[{Vasilyev et~al.(2020)Vasilyev, Dharnidharka, and
  Bohannon}]{vasilyev2020fill}
Oleg Vasilyev, Vedant Dharnidharka, and John Bohannon. 2020.
\newblock Fill in the blanc: Human-free quality estimation of document
  summaries.
\newblock In \emph{Proceedings of the First Workshop on Evaluation and
  Comparison of NLP Systems}, pages 11--20.

\bibitem[{Wolf et~al.(2020)Wolf, Debut, Sanh, Chaumond, Delangue, Moi, Cistac,
  Rault, Louf, Funtowicz, Davison, Shleifer, von Platen, Ma, Jernite, Plu, Xu,
  Le~Scao, Gugger, Drame, Lhoest, and Rush}]{wolf-etal-2020-transformers}
Thomas Wolf, Lysandre Debut, Victor Sanh, Julien Chaumond, Clement Delangue,
  Anthony Moi, Pierric Cistac, Tim Rault, Remi Louf, Morgan Funtowicz, Joe
  Davison, Sam Shleifer, Patrick von Platen, Clara Ma, Yacine Jernite, Julien
  Plu, Canwen Xu, Teven Le~Scao, Sylvain Gugger, Mariama Drame, Quentin Lhoest,
  and Alexander Rush. 2020.
\newblock \href {https://doi.org/10.18653/v1/2020.emnlp-demos.6} {Transformers:
  State-of-the-art natural language processing}.
\newblock In \emph{Proceedings of the 2020 Conference on Empirical Methods in
  Natural Language Processing: System Demonstrations}, pages 38--45, Online.
  Association for Computational Linguistics.

\bibitem[{Yuan et~al.(2021)Yuan, Neubig, and Liu}]{yuan2021bartscore}
Weizhe Yuan, Graham Neubig, and Pengfei Liu. 2021.
\newblock Bartscore: Evaluating generated text as text generation.
\newblock \emph{Advances in Neural Information Processing Systems},
  34:27263--27277.

\bibitem[{Zhang et~al.(2019)Zhang, Kishore, Wu, Weinberger, and
  Artzi}]{zhangbertscore}
Tianyi Zhang, Varsha Kishore, Felix Wu, Kilian~Q Weinberger, and Yoav Artzi.
  2019.
\newblock Bertscore: Evaluating text generation with bert.
\newblock In \emph{International Conference on Learning Representations}.

\bibitem[{Zhang et~al.(2004)Zhang, Vogel, and
  Waibel}]{zhang-etal-2004-interpreting}
Ying Zhang, Stephan Vogel, and Alex Waibel. 2004.
\newblock \href {http://www.lrec-conf.org/proceedings/lrec2004/pdf/755.pdf}
  {Interpreting {BLEU}/{NIST} scores: How much improvement do we need to have a
  better system?}
\newblock In \emph{Proceedings of the Fourth International Conference on
  Language Resources and Evaluation ({LREC}{'}04)}, Lisbon, Portugal. European
  Language Resources Association (ELRA).

\bibitem[{Zhao et~al.(2019)Zhao, Peyrard, Liu, Gao, Meyer, and
  Eger}]{zhao2019moverscore}
Wei Zhao, Maxime Peyrard, Fei Liu, Yang Gao, Christian~M Meyer, and Steffen
  Eger. 2019.
\newblock Moverscore: Text generation evaluating with contextualized embeddings
  and earth mover distance.
\newblock In \emph{Proceedings of the 2019 Conference on Empirical Methods in
  Natural Language Processing (EMNLP)}.

\bibitem[{Zhao et~al.(2023)Zhao, Strube, and Eger}]{zhao-etal-2023-discoscore}
Wei Zhao, Michael Strube, and Steffen Eger. 2023.
\newblock \href {https://aclanthology.org/2023.eacl-main.278} {{D}isco{S}core:
  Evaluating text generation with {BERT} and discourse coherence}.
\newblock In \emph{Proceedings of the 17th Conference of the European Chapter
  of the Association for Computational Linguistics}, pages 3865--3883,
  Dubrovnik, Croatia. Association for Computational Linguistics.

\bibitem[{Zhou et~al.(2022)Zhou, Blodgett, Trischler, Daum{\'e}~III, Suleman,
  and Olteanu}]{zhou-etal-2022-deconstructing}
Kaitlyn Zhou, Su~Lin Blodgett, Adam Trischler, Hal Daum{\'e}~III, Kaheer
  Suleman, and Alexandra Olteanu. 2022.
\newblock \href {https://doi.org/10.18653/v1/2022.naacl-main.24}
  {Deconstructing {NLG} evaluation: Evaluation practices, assumptions, and
  their implications}.
\newblock In \emph{Proceedings of the 2022 Conference of the North American
  Chapter of the Association for Computational Linguistics: Human Language
  Technologies}, pages 314--324, Seattle, United States. Association for
  Computational Linguistics.

\end{thebibliography}
\bibliographystyle{acl_natbib}

\appendix

\end{document}